\documentclass[review]{elsarticle}
\usepackage{amsmath,amssymb,amsfonts}
\usepackage{algorithmic}
\usepackage{graphicx}
\usepackage{textcomp}
\usepackage{hyperref}
\usepackage{pgfplots,lmodern}
\usepackage{tikz}
\def\checkmark{\tikz\fill[scale=0.4](0,.35) -- (.25,0) -- (1,.7) -- (.25,.15) -- cycle;}
\pgfplotsset{compat=newest}

\journal{}

\begin{document}

\begin{frontmatter}

\title{DeepHGNN: Study of Graph Neural Network based Forecasting Methods for Hierarchically Related Multivariate Time Series}

\author[1]{Abishek~Sriramulu \corref{cor1}}%\corref{cor1}
\author[2,3]{Nicolas~Fourrier}
\author[4,1]{Christoph~Bergmeir}

\address{abishek.sriramulu@monash.edu, nfourrier@gmail.com, christoph.bergmeir@monash.edu}
\address[1]{Department of Data Science \& Artificial Intelligence, Monash University, Clayton VIC 3800, Australia}
\address[2]{Being AI, Auckland, New Zealand}
\address[3]{Léonard de Vinci Pôle Universitaire, Research Center, 92 916 Paris La Défense, France}
\address[4]{Department of Computer Science and Artificial Intelligence, University of Granada, Spain}

\cortext[cor1]{Postal Address: Faculty of Information Technology, P.O. Box 63 Monash University, Victoria 3800, Australia. E-mail address: abishek.sriramulu@monash.edu}

\begin{abstract}
Graph Neural Networks (GNN) have gained significant traction in the forecasting domain, especially for their capacity to simultaneously account for intra-series temporal correlations and inter-series relationships. This paper introduces a novel Hierarchical GNN (DeepHGNN) framework, explicitly designed for forecasting in complex hierarchical structures. The uniqueness of DeepHGNN lies in its innovative graph-based hierarchical interpolation and an end-to-end reconciliation mechanism. This approach ensures forecast accuracy and coherence across various hierarchical levels while sharing signals across them, addressing a key challenge in hierarchical forecasting. A critical insight in hierarchical time series is the variance in forecastability across levels, with upper levels typically presenting more predictable components. DeepHGNN capitalizes on this insight by pooling and leveraging knowledge from all hierarchy levels, thereby enhancing the overall forecast accuracy. Our comprehensive evaluation set against several state-of-the-art models confirm the superior performance of DeepHGNN. This research not only demonstrates DeepHGNN’s effectiveness in achieving significantly improved forecast accuracy but also contributes to the understanding of graph-based methods in hierarchical time series forecasting.
\end{abstract}


\end{frontmatter}

\section{Introduction}
Hierarchical forecasting is a method used to forecast data at different levels of hierarchy. This approach considers the interrelationships and dependencies between the various levels of the hierarchy, ensuring that the forecasts at each level are consistent with those at higher levels. For example, suppose that a retail chain has stores in different cities across the country. The demand for a particular product in each store can be forecast, and these forecasts can be aggregated to obtain a regional forecast for each city. The regional forecasts can then be aggregated to obtain a national forecast for the entire organization. Hierarchical forecasting can provide valuable insights into the performance of different parts of an organization, and facilitate better decision-making. It can also help improve inventory planning and management, supply chain management, and financial budgets.

The challenge with hierarchical forecasting is that forecasts must be coherent across the hierarchy. This means that the forecasts at the lower levels of the hierarchy should add up to those at higher levels. For instance, the sum of the demand forecasts for each store in a region should add up to the forecast for that region as a whole. 

Most state-of-the-art hierarchical forecasting methods follow a two-step process:  

1. The forecasting step produces individual forecasts for all or a subset of the time series within the hierarchy.  

2. The reconciliation step ensures the coherency of the forecasts across the hierarchy.

One common approach for achieving coherence in hierarchical forecasting is the top-down approach. In this approach, the forecast at the highest level of the hierarchy is first generated, and then the forecasts at the lower levels are derived by disaggregating the higher level forecasts \cite{nenova2016determining}. This approach is easy to implement and can be useful when forecasting at higher levels of the hierarchy. However, forecasts generated at lower levels of the hierarchy may not be as accurate as those generated using disaggregation methods. Another approach is the bottom-up approach, where forecasts for the lower levels are first generated and then aggregated to obtain higher level forecasts \cite{nenova2016determining}. This approach can be useful when the focus is on forecasting at lower levels of the hierarchy. Recently, optimal reconciliation approaches \cite{hyndman2011optimal} have become popular. They involve optimizing a reconciliation function that ensures that the forecasts generated at each level of the hierarchy are consistent with the forecasts generated at adjacent levels of the hierarchy. The reconciliation function can be formulated as a linear or non-linear optimization problem, and it can be solved using various methods, including linear and quadratic programming. The optimal reconciliation approach has been shown to be effective for improving the accuracy of hierarchical forecasting. The algorithm involves generating forecasts for individual time series at all levels of the hierarchy using a suitable forecasting method and then reconciling these forecasts for coherency \cite{hyndman2011optimal}. These methods are built upon strong assumptions, for example, model forecasts are unbiased, model residuals are jointly covariance stationary, and they are not scalable to increasing volumes of time series in a hierarchy \cite{wickramasuriya2019optimal}. Furthermore, these approaches perform poorly on large data sets because highly disaggregated data tend to have a low signal-to-noise ratio.

Recently, graph neural networks (GNN) have been proven to be able to effectively model both inter-series relationships and intra-series patterns \cite{infosci5,feng2024evhf,xie2019sequential,shin2023performance,gao2023graph}. GNNs can be very helpful in hierarchical forecasting, where the goal is to predict multiple related time series with a hierarchical structure, such as sales data for different products and regions. GNNs are a type of neural network designed to operate on graph structured data, which makes them well-suited for modeling relationships between different time series in a hierarchical structure.

In a hierarchical forecasting problem, the time series can be organized into a tree or a directed acyclic graph (DAG) based on their hierarchy. The nodes in the graph represent the time series and the edges represent the hierarchical relationships between them. GNNs can be used to capture these hierarchical relationships by propagating information through the graph, allowing the model to make better predictions for each time series by considering the information from related time series in the hierarchy. GNNs can be used to model the dependencies between different levels of the hierarchy. For example, the model can learn to propagate information from a higher level time series (such as regional sales data) to a lower level time series (such as sales data for individual stores) to make better predictions. Overall, GNNs offer a powerful tool for modeling the complex relationships between time series in a hierarchical forecasting problem, which can lead to more accurate predictions and better decision making. Building upon this foundation, our study introduces several novel contributions to the field:
\begin{itemize}
    \item We present a Hierarchical Graph Neural Network (DeepHGNN) framework that leverages hierarchical structures for enhanced forecasting accuracy across hierarchical levels, utilizing an end-to-end reconciliation mechanism.
    \item We contribute to the advancement of graph-based methods in hierarchical time series forecasting, showcasing the effective use of hierarchical and graph-based data representations.
    \item Our approach demonstrates the ability of GNNs to outperform several state-of-the-art models in forecasting accuracy through comprehensive evaluations, establishing a new benchmark in the domain.
\end{itemize}

\section{Related Works}
\subsection{GNN for Multivariate Time Series Forecasting}
GNNs offer a promising approach for multivariate time series forecasting because of their ability to capture complex dependencies between variables. GNNs are a type of neural network that operate on graphs, which are mathematical objects that represent relationships between entities. 

\begin{figure}[htb]
  \centering
  \includegraphics[scale= 0.22]{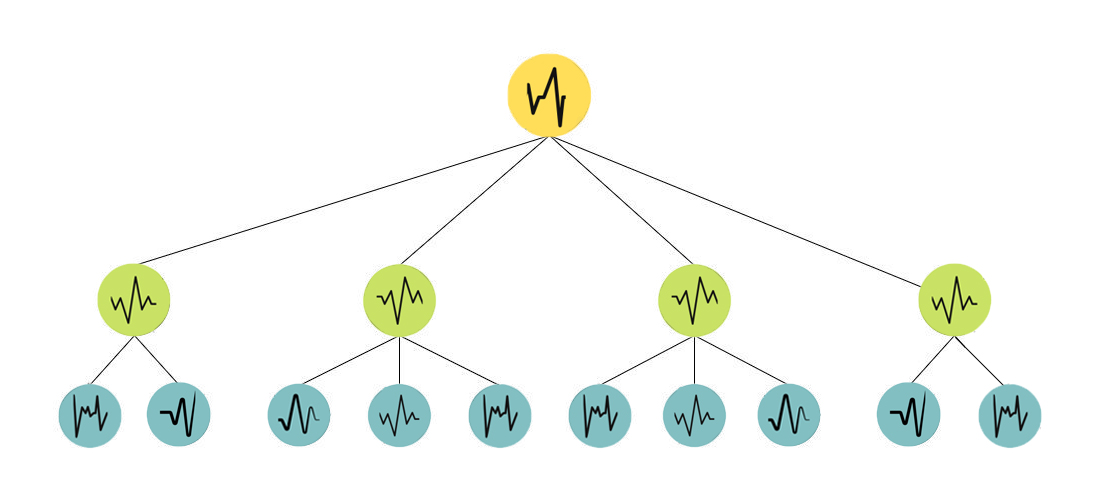}\\
  \caption{Hierarchical data represented in the form of a graph.}\label{Graph_Example}
\end{figure}

In the context of multivariate time series forecasting, entities could be different variables, and the relationships could be dependencies between them over time. Figure~\ref{Graph_Example} shows the graph structure in which each node corresponds to an individual series, and the edges connecting them indicate the relationships between the series.
Traditional forecasting methods often rely on linear or simple non-linear relationships between variables, which may not be sufficient for modeling more complex relationships. Graph neural networks (GNNs) have emerged as a promising approach to overcome this limitation by considering the entire graph structure of the data. GNNs can learn complex relationships by considering the dependencies between different variables, which can be represented as nodes and edges in a graph.

A graph is a data structure that defines dependencies between objects or entities. In a graph, each object or entity is represented by a node and the dependencies between them are represented by edges. The total number of nodes in the graph is denoted by $n$. The graph structure can be defined by an adjacency matrix, represented mathematically as $A \in \mathbb{R}^{n \times n}$. In the adjacency matrix, $A[i,j] > 0$ if there is an edge between nodes $v_i$ and nodes $v_j$, and $A[i,j] = 0$ if there is no edge between nodes $v_i$ and nodes $v_j$. The adjacency matrix captures the dependencies between nodes in the graph.

GNNs are neural networks that operate on graph structured data by propagating information across a graph using message-passing systems. The basic idea of a message-passing system is that each node in the graph aggregates information from its neighbors, processes the information, and sends a message to its neighbors. This process was repeated until a stable state was achieved. The final output of the GNN is obtained by aggregating the information from all the nodes in the graph.

Different types of GNN architectures can be used for multivariate time series forecasting \cite{infosci1, infosci2, infosci3, infosci4}, such as graph convolutional networks (GCNs) and graph attention networks (GATs) which operate on graph structured data by performing convolutional operations on the graph, similar to how traditional convolutional neural networks operate on image data. GCNs use a filter function to combine the features of neighboring nodes and produce a new feature for each node. This process can be repeated multiple times to capture more complex dependencies between nodes in the graph.

GATs use attention mechanisms to weigh the importance of different nodes and edges in a graph, allowing for more flexibility in modeling complex relationships. GATs compute the weight for each edge in the graph based on the features of the nodes connected by the edge. These weights are then used to compute the weighted sum of the features of the neighboring nodes. The attention mechanism allows the model to focus on the most relevant information in the graph for each node, thereby improving the accuracy of the forecasting task.

One of the main advantages of GNNs for time series forecasting is their ability to capture complex relationships between variables. Traditional forecasting methods often assume linear or simple non-linear relationships between variables, which may not be sufficient for modeling more complex relationships. GNNs can learn more complex relationships by considering the entire graph structure of the data. This allows the model to capture the dependencies between different variables, thereby improving the accuracy of the forecasting task. As this area of research continues to grow, GNNs are likely to become increasingly popular and effective in forecasting tasks. 

Various graph neural network models have been proposed for multivariate time series forecasting to capture complex spatio-temporal dependencies, including DCRNN \cite{li_diffusion2018} which combines graph convolution and Recurrent Neural Network (RNN) encoder-decoder, MTGNN \cite{wu2020connecting} with graph convolution and dilated temporal convolution modules, ADLGNN \cite{sriramulu2023adaptive} using adaptive dependency learning attention, MTGODE \cite{liu2023graph} modeling evolution via spatial and temporal ODEs, ASTGCN \cite{bai2021a3t} with a Gated Recurrent Unit (GRU), graph convolution and attention, and SpecTGNN \cite{jin2023expressive} which proves spectral-temporal GNNs as a universal approximator. These methods aim to effectively exploit spatial correlations, long and short term temporal patterns, and dynamical evolution by combining techniques such as graph operations, attention, temporal convolutions, and neural ODEs.

\subsection{Hierarchical forecasting}

Hierarchical time series forecasting involves forecasting a large number of related time series that are organized into a hierarchical structure. Hierarchical structures can be found in many domains, including sales forecasting, demand forecasting, and supply chain management. In such domains, the forecasts of individual time series at the lowest level of the hierarchy are aggregated to generate forecasts at higher levels. Hierarchical forecasting is a powerful technique for forecasting large numbers of related time series because it allows for the exploitation of information at all levels of the hierarchy. Moreover, hierarchical forecasting can provide more accurate forecasts for individual time series by incorporating information from the related time series.

Hierarchical time series forecasting is often performed as a two-step process, where first forecasts for all time series on all levels are produced, and afterwards, they are modified to coherently add up in a subsequent reconciliation step \cite{hollyman2021understanding}. More recently, end-to-end systems have been proposed, where the model jointly optimizes both forecasting and reconciliation and therefore promises to achieve better accuracy. In this study, we focus on such end-to-end solutions, discuss them and compare them with DeepHGNN. Table \ref{sota-table} shows the characteristics of existing end-to-end hierarchical forecasting solutions.

\subsubsection{SHARQ} \label{SHARQ} 
\citet{han2021simultaneously} proposed a method for capturing the mutual relationships among related series from adjacent aggregation levels. This method involves optimizing the regularized loss function. The model learns the weights for each vertex in the summing matrix which acts as a regularization term to enforce the coherency of forecasts across hierarchies. Even though this approach optimizes coherency, it does not guarantee coherent forecasts.

The time series value at time $t$ for node $v_i$ in the hierarchical structure is denoted by $x_{v_i}(t)$. The parent-child relationships between nodes are represented using signed edges $e_{i,j}$, where $e_{i,j} = 1$ if $i$ is the parent of $j$, and $e_{i,j} = -1$ if $j$ is the parent of $i$.

For each node $i$, a forecasting model $g_i$ is trained that takes the past data $X_i$ and model parameters $\theta_i$ as input, and outputs a prediction for the future values $Y_i$. Flexibility is allowed in the choice of $g_i$ - which can be a simple linear model or complex deep neural network based on the characteristics of the dataset.

The forecast model $g_i$ minimizes the overall constrained objective function $L_c$ which is a combination of the data fit loss $L$ and reconciliation regularization:

$$ L(g_i(X_i,\theta_i), Y_i) =  \sum_{m=1}^M L(g_i(X_i^m, \theta_i), Y_i^m) $$
where $g_i(X_i,\theta_i)$ and $Y_i$ represents the model predictions and true values  over the $m$ training samples.

Coherence across the hierarchical structure is enforced by adding a regularization term that minimizes the difference between the forecast of a parent node $i$ and the aggregate forecast from its child nodes $j$. 
The strength of this coherence constraint is controlled by the regularization weight $\lambda_i$. Intuitively, this term pulls the parent forecast towards the cumulative forecast from children. The overall constrained objective function combines the data fit loss and reconciliation regularization:

$$ L_c(g_i, Y_i, g_j) = L(g_i, Y_i) + \lambda_i \left| g_i - \sum_{j} e_{i,j} g_j \right|^2 $$

Optimizing this joint loss function enables the learning of model parameters $\theta_i$ that balance the accuracy of individual time series with the coherence across the hierarchy. Hierarchical regularization allows the dependencies between the aggregation levels to be learned by the model.

\subsubsection{HIRED} \label{HIRED}
\citet{paria2021hierarchically} introduced a new approach for hierarchical time series forecasting that consists of two main components: a time-varying autoregressive (TVAR) model and a basis decomposition (BD) model that can simultaneously forecast a large number of hierarchically structured time series while exploiting correlations along the hierarchy tree to improve accuracy. 
$$ \hat{y}^{(i)}_{t+1} =  \{\hat{y}^{TVAR(i)}_{t+1}\} + \{\hat{y}^{BD(i)}_{t+1}\}  $$
The TVAR model uses a linear autoregressive formulation with time-varying coefficients:
$$ \hat{y}^{TVAR(i)}_{t+1} = { y^{(i)}_{t-H:t-1}, a(X_{t-H:t}, Z_{t-H:t}) }  $$
where, $\hat{y}^{(i)}_{t+1}$ is the one-step forecast for time series $i$ at time $t+1$, $y^{(i)}_{t-H:t-1}$ is the $H$-step history of series $i$, $X_{t-H:t}$ are global input features, $Z_{t-H:t}$ are representative time series capturing global temporal patterns, and $a(.)$ is a learned vector function producing AR coefficients as the output.

The AR coefficients are shared across all the time series. However, the coefficients can change over time based on the input features and global patterns, providing flexibility. The BD model decomposes each time series into a linear combination of global basis functions $b(.)$ with the time series specific weights $\theta_i$.
$$  \hat{y}^{BD(i)}_{t+1} = { \theta_i, b(X{t-H:t}, Z_{t-H:t}) }  $$
The basis functions $b(.)$ and weights $\theta_i$ are learned jointly. To encourage coherence, the weights are regularized to satisfy the hierarchy constraints. The TVAR part models linear autocorrelations and can adapt over time, whereas the BD part captures global temporal effects and cross-series correlations through a shared basis. The model is trained end-to-end to minimize the mean absolute error loss along with regularization of the BD weights to enforce coherence. Crucially, the model can be trained with mini-batches without requiring the entire batch to contain all the time series, enabling scaling to large datasets. Although this regularization promotes coherence, it does not ensure perfect coherency. There may still be some degree of variability in the forecasts.

\subsubsection{PROFHIT} \label{PROFHIT} 
\citet{kamarthi2022profhit} proposed a probabilistic neural network model that first learns the forecast distribution using historical data of each time series. Using this as a prior, a refined set of distributions is produced. This enables the modeling of information from individual time series along with their hierarchical information. Furthermore, this model provides flexibility in incorporating various levels of consistency towards the hierarchical constraints. 
The model consists of two modules:

\begin{itemize}

\item TSFNP Module: TSFNP encodes history $y_i^{(1:t)}$ into a latent embedding and models a Gaussian distribution. This predicts the raw forecast distribution $p(y_i^{(t+\tau)}|D_t)$ for each time series $i$. 

It has 3 components:  
\begin{itemize}
\item{Probabilistic Neural Encoder (PNE):} Encodes the history $y_i^{(1:t)}$ into a stochastic latent variable $u_i$ using GRU and self-attention.  

\item{Stochastic Data Correlation Graph (SDCG):} Builds correlations between $u_i$ and encodings $u_j$ of other time series. The aggregates correlate $u_j$ within a local latent variable $z_i$.  

\item{Predictive Distribution Decoder (PDD):} Predicts mean $\hat{\mu}_i$ and variance $\hat{\sigma}_i$ of the Gaussian raw forecast distribution from $u_i$, $z_i$ and global correlations.
\end{itemize}

\item Refinement module: Raw forecast distributions are refined using a novel method that combines information across all time series. The refined mean $\mu_i$ is a weighted combination of the raw mean $\hat{\mu}_i$ and the raw means of other related time series $\hat{\mu}_j$. The refined variance $\sigma_i$ depends on the raw variances $\hat{\sigma}_j$.
\end{itemize}

Distributional coherency loss is used to regularize the refined distributions towards coherence with the hierarchy. It measures the divergence between $p(y_i^{(t+\tau)}|D_t)$ and the distribution of the aggregated child nodes $\sum_{j \in C_i} \phi_{ij} y_j^{(t+\tau)}$. 
Although this is an end-to-end model, it requires pretraining of the TSFNP to predict raw distributions before end-to-end training.

\subsubsection{HierE2E}  \label{HierE2E}
\citet{rangapuram2021end} proposed an end-to-end deep neural network based hierarchical forecasting model, that guarantees coherency, unlike models that use regularization techniques for enforcing hierarchical constraints.
DeepVAR, a multivariate forecasting model is used to generate a joint predictive distribution over the future time steps.

Let $y_t \in \mathbb{R}^n$ denote the hierarchical time series at time step $t$ with $y_{t,i}$ is the $i^{th}$ univariate series. The model is:

$$ \Theta_t = \Psi(x_t, y_{t-1}, h_{t-1}; \Phi)    $$
$$ p(y_t; \Theta_t) = \mathcal{N}(y_t; \mu_t, \Sigma_t) $$
Here, $\Theta_t = \{ \mu_t, \Sigma_t \}$ are the distribution parameters predicted using network $\Psi$ with parameters $\Phi$. $x_t$ and $h_{t-1}$ are the input features and previous hidden state, respectively. Unlike independent forecasting, the mean $\mu_{t,i}$ and variance $\Sigma_{t,ii}$ for each series depend on the lags $y_{t-1}$ of all the series.

The overall model comprising the forecasting network and differentiable sampling and projection steps, can be trained end-to-end using any loss function. The coherent samples $\hat{y}_t$ directly provide reconciled probabilistic forecasts.

\subsubsection{DPMN} \label{DPMN} 
\citet{olivares2021probabilistic} proposed a Deep Poisson Mixture Network (DPMN), which is a combination of neural networks and statistical models. First, the predictive distribution is modeled as a finite mixture of Poisson random variables which is analogous to a kernel density with Poisson kernels. This distribution was used to model the probability distribution using a neural network (NN).

The key idea is to model the joint distribution of the bottom level time series by using a finite mixture of Poisson distributions. This allows the method to flexibly model correlations across time series while guaranteeing hierarchical coherence in predictive distributions.

\begin{align*}
\text{Let } \Lambda_{bkt} &= \lambda_{[b][k][t+1:t+h]}, \\
Y_{bt} &= y_{[b][t+1:t+h]}, \\
\text{PT}_{\beta,\tau,k} &= \text{Poisson}\left(y_{\beta,\tau} | \lambda_{\beta,k,\tau}\right), \\
\hat{P}(Y_{bt} | \Lambda_{bkt}) &= \sum_{k=1}^{K} w_{k} \times \prod_{(\beta,\tau) \in [b][t+1:t+h]} \text{PT}_{\beta,\tau,k}
\end{align*}

where $y_{[b][t+1:t+h]}$ is the future bottom level multivariate time series, $\lambda_{[b][k][t+1:t+h]}$ is the poisson rate parameterized by the neural network, $w_{k}$ are the mixture weight, and $K$ is the number of mixture components. The Poisson rates $\lambda_{[b][k][t+1:t+h]}$ and weights $w_{k}$ is learned by maximizing the composite likelihood, which breaks up the high dimensional joint distribution into lower dimensional spaces for computational efficiency. The architecture uses dilated temporal convolutions to encode past observations and future co-variates into context vectors $c^{(h)}_{t}$, which are fed to MLP decoders to output poisson parameters.
Forecasts at the aggregate levels were computed by adding Poisson rates:
$$
\lambda_{[a][k],\tau} = A_{[a][b]} \lambda_{[b][k],\tau}
$$

where $A_{[a][b]}$ is the aggregation matrix. This guarantees hierarchical coherence by construction.

 \begin{center}
\begin{table}[!htb]
\label{sota-table}
 \caption{Comparison of state-of-the-art hierarchical end-to-end forecasting methods}
\begin{tabular}{|c|c|c|c|} 
 \hline
 Model & Guaranteed Coherence & End-to-End Model & Update-able \\ 
  \hline
 SHARQ & - & \checkmark & - \\ 
 HIRED & - & \checkmark & -  \\ 
 PROFHIT & - & \checkmark & -  \\ 
 HierE2E  & \checkmark & \checkmark & -  \\ 
 DPMN & \checkmark & \checkmark & \checkmark  \\
 DeepHGNN (Ours)& \checkmark & \checkmark & \checkmark  \\
 \hline
\end{tabular}
\end{table}
\end{center}

\section{Proposed Method Design \& Framework}

The Hierarchical Graph Neural Network (DeepHGNN) is a promising end-to-end hierarchical forecasting design that aims to improve the accuracy of hierarchical time series forecasting by augmenting the target series with more information from its hierarchy. 

The DeepHGNN model was formulated as follows: Let $\boldsymbol{h}_{t}=\boldsymbol{S} \boldsymbol{b}_{t} \in \mathbb{R}^{m}$ be an $m$-vector containing the observations at time $t$, where $\boldsymbol{S} \in\{0,1\}^{m \times n}$ is the summing matrix that aggregates the series at the bottom level, and $\boldsymbol{b}_{t} \in \mathbb{R}^{n}$ is the $n$-vector of the bottom level series at time $t$, which is used to form $\boldsymbol{h}_{t} \in \mathbb{R}^{a}$. Additionally, let $\boldsymbol{f}_{t}^{bi} \in \mathbb{R}^{k_{bi}}$ be a $k_{bi}$-vector containing the features from the bottom level series $bi$ at time $t$ for $bi=1,2, \ldots, n$ and $\boldsymbol{f}_{t}^{hi} \in \mathbb{R}^{k_{hi}}$ be a $k_{hi}$-vector containing the features from aggregated series $hi$ at time $t$ for $hi=1,2, \ldots, a$. The model also accommodates co-variates, external regressors, and changes in hierarchical structures. The DeepHGNN model exhibits a better modeling capability by propagating information across a hierarchical structure using GNNs. Specifically, GNNs allow the model to learn the relationships between the different levels of the hierarchy and the features associated with each level. This enables the model to capture the dependencies between the bottom level series and the higher level series and to leverage the information from all levels to improve the forecast accuracy.

Traditional hierarchical forecasting models assume a fixed hierarchy, which may not be appropriate for datasets in which the underlying structure changes over time. DeepHGNN can accommodate such changes by adapting adjacency matrix $\boldsymbol{A}$ during model training to capture the evolving hierarchy. This adaptability is valuable in dynamic environments, where the relationships between series may evolve.

\begin{figure}[!tbp]
\vspace{-3cm}
  \centering
  \includegraphics[width=0.4\textwidth]{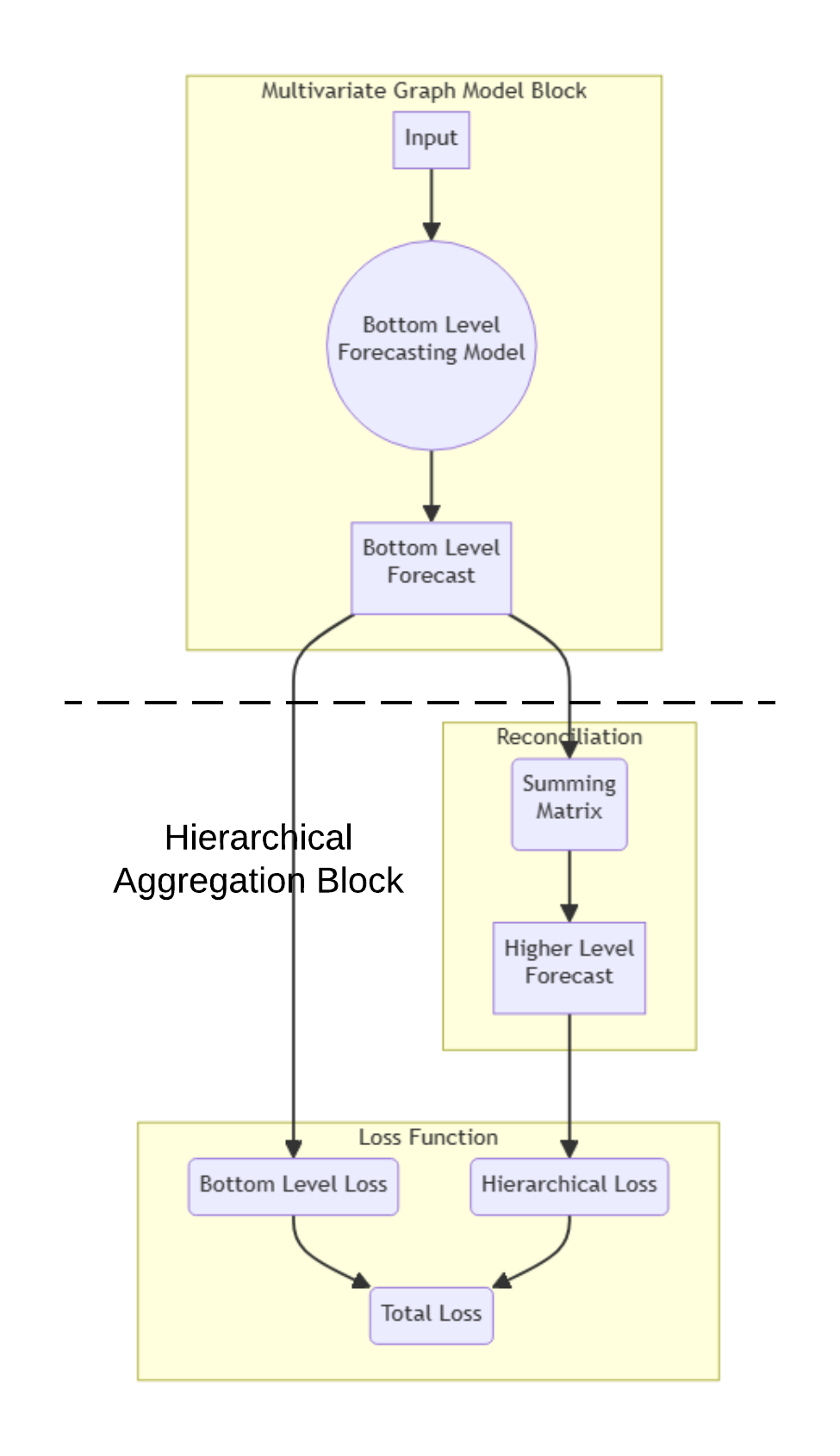}
  \caption{DeepHGNN Framework}
  \label{fig:DeepHGNN framework}
\end{figure}

As shown in Figure \ref{fig:DeepHGNN framework}, DeepHGNN is designed in two blocks: 
\begin{itemize}
    \item Multivariate Graph Model (MGM) Block: This block can be any multivariate forecasting GNN model that models spatio-temporal patterns.
    \item Hierarchical Aggregation Block: This block takes in forecasts of a multivariate time series in the shape of (number of nodes, forecast horizon, number of output features) and outputs hierarchically reconciled forecasts.
\end{itemize}

\begin{figure}[!tbp]
  \centering
  \includegraphics[width=\linewidth]{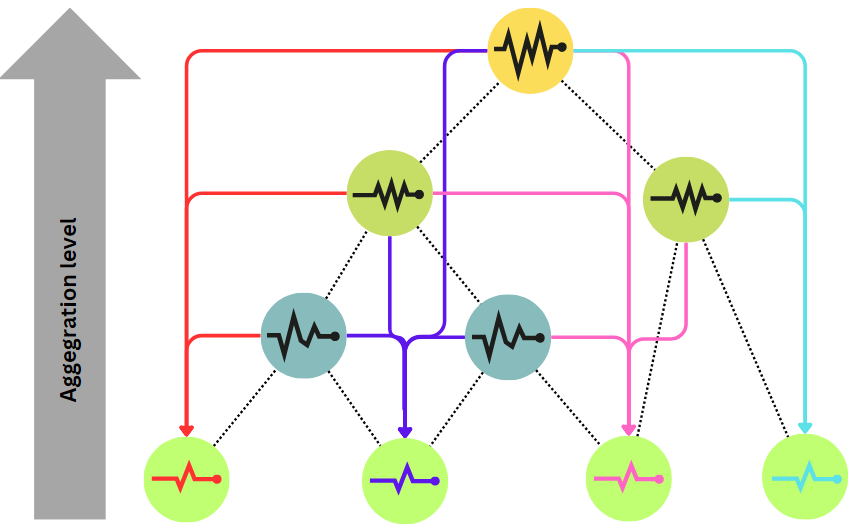}
  \caption{Hierarchical Information propagation (The arrows represent information flow and dotted lines represent hierarchical connection)}
  \label{fig:info prop DeepHGNN}
\end{figure}

\subsection{Multivariate Graph Model Block}
The MGM block generates forecasts for bottom level series. One of its key strengths in the utilization of a graph model is that it enables the incorporation of information from both parents and ancestor series in the hierarchy while learning to model the target series. The use of a graph model is significant because it allows the forecasting process to consider the relationships and dependencies between different series within a hierarchy. In a hierarchical dataset, we typically have a series at different levels, with some series influencing or being influenced by others. For example, in the retail context shown in Figure \ref{fig:info prop DeepHGNN}, store sales can be influenced by regional trends, which in turn can be influenced by national conditions. 
When the MGM block generates forecasts, it leverages the current state of its ancestors. This implies that unlike traditional hierarchical forecasting models, this model does not treat each series in isolation. Instead, it considers the information from a higher level (ancestor) series to make more reliable predictions for the target series. The signal-to-noise ratio improves as we move up the hierarchy. A graph model can leverage this signal rich information to enhance forecast accuracy. 

In the field of Graph Neural Networks, particularly when dealing with multivariate forecasting, various approaches have been discussed in the literature. It is important to explore and utilize these different possibilities to determine which ones work best in the MGM block, which is why we present the methodology of these approaches in detail in the following.

\subsubsection{DCRNN} \label{DCRNN}

\citet{li_diffusion2018} proposed a graph diffusion based convolutional recurrent neural network (DCRNN) for traffic forecasting. The DCRNN is based on the LSTM model but has several key differences. First, the DCRNN uses a CNN to extract features from time series data which allows the DCRNN to learn long-range dependencies in the data, which is important for forecasting. Second, the DCRNN uses an RNN to model the temporal dynamics of data. This allows the DCRNN to learn how data changes over time.

\sloppy
This model forecasts $X^{(t+1)},\dots,X^{(t+T)}$ given historical observations $X^{(t-T')},\dots,X^t$ from $N$ sensors on a network. The network is modeled as a weighted directed graph $G=(V,E,W)$ where $V$ denotes the set of nodes (sensors), $E$ denotes the set of edges, and $W$ denotes the weighted adjacency matrix encoding proximity between sensors based on the network distance represented as graph signals $X \in \mathbb{R}^{N\times P}$ on $G$.

The spatial dependencies were modeled by the diffusion process on $G$. This diffusion process is characterized as a random walk on $G$ with restart probability $\alpha$. The stationary distribution $P$ of this random walk captures the correlations between the nearby nodes on the graph. Specifically, the $i$'th row $P_{i,:}$ represents the proximity of all nodes to node $i$. $P$ is provided in terms of the powers of the graph random walk transition matrix:

$$ P = \sum_{k=0}^{\infty} \alpha(1-\alpha)^k (D^{-1} W)^k $$

Here, $D^{-1}$ is the diagonal out-degree matrix and $k$ is the diffusion step.
Based on this diffusion process, a novel diffusion convolution operation was proposed. This convolution between a graph signal $X$ and filter $f_\theta$ is defined as a weighted sum of the diffusion outputs using the powers of the random walk transition matrix and its transpose:

$$ X \star G f_\theta = \sum_{k=0}^{K-1} \theta_{k,1} (D^{-1} W)^k X + \theta_{k,2} (D^{-1} W^T)^k X $$

where $\theta \in \mathbb{R}^{K\times 2}$ are filter parameters.
This captures the localized patterns diffusing over the graph, analogous to regular convolutional filters. Regular matrix multiplications in the GRU are replaced with the proposed graph diffusion convolutions. This enables the modeling of spatio-temporal correlations. An encoder-decoder architecture with scheduled sampling was used for multi-step forecasting. The encoder uses in historical sequences, whereas the decoder initialized with the encoder state produces predictions. The model was trained by maximizing the likelihood of generating future sequences using backpropagation over time. 

\subsubsection{MTGNN} \label{MTGNN}
 
\citet{wu2020connecting} proposed a time series forecasting model based on graph neural networks and dilated temporal convolutions. The key components of the MTGNN are as follows:

\begin{itemize}

\item \textbf{Graph Convolution Module}: Exploits the spatial dependencies among variables using a novel mix-hop propagation layer for graph convolution over the learned adjacency matrix. It avoids over-smoothing and retains locality. This module performs graph convolutions over the adjacency matrix $A$ to aggregate spatial dependencies. It consists of two novel mix-hop propagation layers that process the inflow and outflow information separately.

The mix-hop propagation layer has two steps:  \\
\textit{Information Propagation}: $$H^{(k)} = \beta H_{in} + (1-\beta) \tilde{A} H^{(k-1)}$$ where $\tilde{A} = \tilde{D}^{-1}(A + I)$, $\tilde{D}_{ii} = 1 + \sum_j A_{ij}$. This propagates information from the input hidden state ($H_{in}$) to k-hop neighbors while retaining a ratio $\beta$ of the root node's state to avoid over-smoothing. \\
\textit{Information Selection}: $$H_{out} = \sum_{k=0}^K H^{(k)} W^{(k)}$$
where $W^{(k)}$ is a learnable parameter that selects the useful features from each propagation step. Mix-hop propagation can represent the interactions between consecutive hops and avoid the averaging effects of standard graph convolutions. The inflow and outflow mix-hop layers capture both incoming and outgoing spatial dependencies.  

\item  \textbf{Temporal Convolution Module}: Captures temporal patterns using dilated convolutional layers with multiple kernel sizes. This module applies 1D dilated convolutional layers to extract the temporal features.   
It consists of two components: \\
\textit{Dilated inception layer}: Four parallel dilated 1D convolutions with kernel sizes of 2, 3, 6 and 7 to capture multi-scale temporal patterns. \\
\textit{Dilation schedule}: Increases the dilation exponent $q$ at each layer to exponentially expand the receptive field. If layer $i$ has dilation $d_i$, then $d_{i+1} = d_i^q$.  \\
The inception layers can represent temporal patterns at different scales, whereas the dilation schedule allows exponentially larger receptive fields with fewer layers. Together, they effectively model both short and long sequences.

\item \textbf{Learning Algorithm}: Uses curriculum learning and sub-graph training to find better optima and handle large graphs.   \\
\textit{Sub-graph Training}: Randomly split nodes into groups during each training iteration to reduce memory and computational complexity.   \\
\textit{Curriculum Learning}: Gradually increase the sequence length for prediction from 1 step to Q steps. This prevents easier short term predictions from dominating optimization.

\end{itemize}

The MTGNN model integrates the spatial and temporal modules into an end-to-end architecture. The major drawback of this model is that the estimated forecast may not always be optimal if the variables have a bi-directional causal influence.

\subsubsection{ADLGNN} \label{ADLGNN-oursc2}

\citet{sriramulu2023adaptive} proposed a solution to reduce the runtime complexity of large GNN models by using statistical methods to identify the neighbors holding more valuable information and applying attention to not only reduce the number of neighbors but also to avoid negative learning. The ADLGNN architecture comprises of two main components: a GNN and an adaptive dependency learning module. The GNN is used to learn intra-series dependencies, whereas the adaptive dependency learning module is used to learn inter-series dependencies. The adaptive dependency learning module is based on a graph attention network (GAT) mechanism. The adaptive dependency learning module is used to learn the attention weights between the time series in the multivariate time series dataset. Attention weights are used to determine the importance of each time series for predicting the next observation in the target time series. A modified version of Spatio-Temporal Attention is used to capture spatio-temporal patterns. This attention mechanism transforms the input using causal convolutions to obtain queries, keys and values that attend to both the spatial and temporal dimensions simultaneously.
The method is similar to standard self-attention.

$$
Attention(q,k,v) = softmax(\frac{qk^T}{\sqrt{d_k}})v
$$

where $q,k,v$ are obtained by convolving the inputs. Attention scores were fed through skip connections to the output module.

\subsubsection{MTGODE} \label{MTGODE}

\citet{jin2022multivariate} proposed a neural graph ordinary differential equation (ODE) that unifies both spatial and temporal message passing, allowing deeper graph propagation and fine-grained temporal information aggregation. MTGODE introduces a novel approach to multivariate time series forecasting that uses dynamic graph neural ODEs. The proposed method can capture the complex temporal dependencies between multiple time series. The ODEs were used to model the evolution of the graph over time, and the GNN was used to learn the relationships between the nodes in the graph. 
The proposed model consists of two ODEs:  

\textit{Spatial ODE} that propagates information along the learned graph to model spatial dependencies:
$$\frac{dH_G(t)}{dt} = (\hat{A} - I_N) H_G(t)$$
where $H_G(t)$ is the latent representation capturing the spatial dependencies at time $t$, $\hat{A}$ is the normalized adjacency matrix of the learned graph, and $I_N$ is the identity matrix. This allows for the continuous propagation of information across the graph, avoiding the over-smoothing problem in traditional GNNs.  

\textit{Temporal ODE} that aggregates information over time:
$$\frac{dH_T(t)}{dt} = P(TCN(H_T(t), t), R)$$
where $H_T(t)$ is the latent representation capturing temporal patterns at time $t$, $TCN$ is a temporal convolution module, $P$ is a padding function, and $R$ is the receptive field size of the model. This allows the continuous-time modeling of temporal patterns, avoiding issues with discrete models. This combination makes the ODE dynamic, whereby the spatial ODE is solved as an inner step within each evaluation of the temporal ODE.

\subsubsection{ASTGCN} \label{ASTGCN}

\citet{bai2021a3t} proposed Attention Temporal Graph Convolutional Network for Forecasting (ASTGCN). This is a novel deep learning model that has been proposed to simultaneously capture global temporal dynamics and spatial correlations for forecasting. The model is composed of three main components: a gated recurrent unit (GRU) layer, GCN layer, and attention mechanism for adjusting the importance of different time points. 
The graph convolutional layer is used to capture the spatial dependencies:
$$H^{(l+1)} = \sigma(\tilde{D}^{-\frac{1}{2}} \tilde{A} \tilde{D}^{-\frac{1}{2}} H^{(l)} \Theta^{(l)})$$
where $H^{(l)}$ is the output of layer $l$, $\tilde{A} = A + I$ is the adjacency matrix with self-connections, $\tilde{D}_{ii} = \sum_j \tilde{A}_{ij}$ is the diagonal degree matrix, and $\Theta^{(l)}$ is the layer weight.  

The GRU layer is used to capture local temporal dynamics:
$$  u_t = \sigma(W_u * [X_t, h_{t-1}] + b_u)  $$
$$ r_t = \sigma(W_r * [X_t, h_{t-1}] + b_r)   $$
$$ c_t = tanh(W_c * [X_t, (r_t * h_{t-1})] + b_c)   $$
$$ h_t = u_t * h_{t-1} + (1 - u_t) * c_t  $$
where $u_t, r_t$ are the update and reset gates, $c_t$ is the candidate memory, $h_t$ is the GRU output, and $X_t$ is the input at time $t$.

The attention mechanism captures the global temporal dynamics. It is used to adjust the importance of different time points by assigning higher weights to the more important time points.
$$ e_i = W^{(2)}(W^{(1)}H + b^{(1)}) + b^{(2)} $$
$$ \alpha_i = \frac{exp(e_i)}{\sum_{k=1}^n exp(e_k)} $$
$$ C_t = \sum_{i=1}^n \alpha_i * h_i $$
where $H = {h_1, ..., h_n}$ are the GRU outputs, $e_i$ is the attention score, $\alpha_i$ is the normalized attention weight, and $C_t$ is the context vector.

\subsubsection{GRAMODE} \label{GRAMODE}

\citet{liu2023graph} proposed graph based multi-ODE neural networks (GRAM-ODE) for capturing complex spatio-temporal patterns. The GRAM-ODE framework contains two input streams for the connection map and DTW graph, three parallel channels per graph with two GRAM-ODE layers and an attention module (AM) to aggregate the final features.

\begin{itemize}
\item \textbf{GRAM-ODE Layer} has a multi-ODE GNN block between two temporal convolutional networks (TCNs). The multi-ODE block is responsible for: \textit{message passing} for global, local, and edge ODE modules; \textit{message filtering} to limit divergence; \textit{aggregation} using gated operations; and \textit{update} via residual connections.
The three message-passing schemes are expressed as
$$ GM = \text{GlobalMessagePassing}(H_g(0), \hat{A}, T_g, W)$$ 
$$ = H_g(0) + \int_{0}^{t} f_g(H_g(\tau), \hat{A}, T_g, W) d\tau $$ where  $$f_g = H_g(t) \times_2 (\hat{A} - I) + ((S(T_g) - I)H_g^T(t))^T + H_g(t) \times_4 (W-I) $$
$$LM = \text{LocalMessagePassing}(ATT, H_l(0), \hat{A}, T_l, W) $$
$$ EM = \text{EdgeMessagePassing}(H_e(0), \hat{A}, T_e)  $$ where $$f_e = H_e(t) \times_2 (\hat{A} - I) + H_e(t)(S(T_e) - I) $$
Shared weights $T_g, T_e$ were used between the global and edge ODEs. The global message-passing scheme models long term temporal patterns whereas the local message-passing and edge message-passing schemes model short term temporal patterns and edge correlations. Message filtering performs a two-way clipping operation that restricts the divergence between local and global modules. The gated aggregation scheme combines the three ODE module outputs, as follows:
$$H' =\text{Aggregation}(GM, LM, EM) $$
$$ =\frac{1}{2K}\sum_{m=1}^{K}\sum_{n\neq m}p_m \odot \text{softmax}(p_n)$$
where $p_m \in {GM, LM, EM}$. This enables selection and non-linear fusion.

\item \textbf{Temporal Convolutional Networks (TCNs)} were used before and after the ODE block to model temporal patterns to capture informative temporal representations. The TCN module applies 1D convolutional operations along the time dimension to capture the temporal dependencies in the data. Dilated convolutions are used to expand the receptive field and to aggregate information from broader contexts. The TCN contains a stack of convolutional layers. For the $l$-th layer:
$$H_t^{TCN(l)} = Sigmoid(W_l *{d_l} H_{t-1}^{TCN(l-1)})$$
where $H_t^{TCN(l)} \in \mathbb{R}^{B \times N \times L \times C_l}$ is the latent output, $B$ is the batch size, $N$ is the number of nodes, $L$ is the  input length, $C_l$ is the number of channels at layer $l$, $W_l \in \mathbb{R}^{k \times C_{l-1} \times C_l}$ is the 1D convolution kernel, $k$ is the kernel size, and $d_l = 2^{l-1}$ is the exponential dilation rate. Dilated convolution enables an exponentially increasing receptive field without a loss of resolution or coverage. Stacking the TCN layers allows the modeling of long-range temporal dependencies. 

\item \textbf{Attention Module (AM)}
The attention module replaces regular fully connected layers to aggregate features across the two graph streams and multiple channels in a more effective way. It uses concatenated embeddings $X \in \mathbb{R}^{B \times N \times L \times C'}$ as input, where $C'$ is the concatenated dimension. The attention mechanism allows the identification of correlations between the input streams and channels that are most relevant to the forecasting task. This provides more expressive feature aggregation than typical approaches such as pooling. AM adaptively fuses information from diverse inputs using data-driven attention scores, enabling the selection of the most informative features.
\end{itemize}

\subsubsection{SpecTGNN} \label{SpecTGNN}

\citet{jin2023expressive} proved that the spectral-temporal graph neural network (SpecTGNN) can be a universal approximators for multivariate time series forecasting under mild assumptions. The design principles outlined in this paper describe how to construct provably expressive SpecTGNN models using orthogonal bases and individual spectral filters. The proposed model, Temporal Graph GegenConv (TGC), validates these theories, showing strong performance with a simple linear architecture. The TGC architecture stacks multiple blocks, with each block applying graph convolution (Gegenbauer basis), coarse + fine-grained temporal frequency-domain filtering, and residual connection. For graph convolutions, TGC uses the Gegenbauer polynomial basis because of its desirable properties such as fast convergence and generalization of Chebyshev polynomials to more flexible weight functions and simplicity compared to other orthogonal bases such as Jacobi. The Gegenbauer polynomials are defined recursively as $$P_k^\alpha(x) = \frac{1}{k}[2x(k + \alpha - 1)P_{k-1}^\alpha(x) - (k + 2\alpha - 2)P_{k-2}^\alpha(x)]$$
with $P_0^\alpha(x) = 1$ and $P_1^\alpha(x) = 2\alpha x$. They are orthogonal over $[-1, 1]$ with respect to the weight function $(1 - x^2)^{\alpha - 1/2}$. The graph convolution operation is as follows: $$g_\theta(\hat{L}) \star X_t = \sum_{k=0}^K \theta_k P_k^\alpha(\hat{A}) X_t$$ where $\hat{A}$ is the normalized adjacency matrix.
TGC uses two simple temporal frequency-domain filtering steps:
\begin{itemize}
\item Coarse-grained: Apply  Discrete Fourier Transform (DFT), select/filter components, and inverse DFT.
\item Fine-grained: Decompose the input into trends and details, filter separately, and recombine.
\end{itemize}
This fine-grained filtering helps capture different frequencies in the input signal.

\subsection{Hierarchical Aggregation Block}

The Hierarchical Aggregation Block performs bottom-up reconciliation combining low level forecasts to generate high level forecasts in a hierarchical time series. The steps are:

\begin{enumerate}
\item The output bottom level time series $\boldsymbol{b}_{t} \in \mathbb{R}^{n}$ from the MGM block with input features $\boldsymbol{f}_{t}^{bi} \in \mathbb{R}^{k_{bi}}$. Let the outputs be $\hat{\boldsymbol{b}}_{t}$.

\item Aggregate the bottom level forecasts using the summing matrix $\boldsymbol{S} \in {0,1}^{m \times n}$ to obtain forecasts for higher levels:   
$$\hat{\boldsymbol{h}}_{t} = \boldsymbol{S}\hat{\boldsymbol{b}}_{t}$$

\item Repeat step 2 recursively to forecast all higher level aggregated series $\hat{\boldsymbol{h}}_{t} \in \mathbb{R}^{m}$ using features $\boldsymbol{f}_{t}^{hi} \in \mathbb{R}^{k_{hi}}$ until the top level forecast is obtained.
\end{enumerate}

The loss function for optimizing the bottom level forecasting model can be defined as:

$$L(\theta) = \sum_{t \in T_{0}} \ell(\boldsymbol{b}_{t}, \hat{\boldsymbol{b}}_{t}(\theta)) + \lambda \sum_{t \in T_{0}} \ell(\boldsymbol{h}_{t}, \hat{\boldsymbol{h}}_{t}(\theta))$$
where $T_0$ is the training set, $\ell()$ is the loss function (e.g., MSE), $\theta$ is the model parameter, $\boldsymbol{b}_{t}$ is the actual bottom level series, $\hat{\boldsymbol{b}}_{t}(\theta)$ is the forecast bottom level series, $\boldsymbol{h}_{t}$ is the actual aggregated series, $\hat{\boldsymbol{h}}_{t}(\theta)$ is the reconciled forecast aggregated series, and $\lambda$ controls the weight assigned to the total aggregated loss.

Using a hierarchical loss function that incorporates errors from both bottom level forecasts and reconciled bottom-up forecasts provides several key advantages for optimizing forecasting models in a hierarchical time series setup. First, it provides a more comprehensive global loss by considering errors made at multiple levels in the hierarchy instead of just one level. This prevents the model from overfitting to a particular level and helps improve the predictive accuracy across both the granular and aggregated series. Second, reconciliation acts as a regularization mechanism that imposes logical consistency between hierarchical levels. By training the model to minimize reconciliation errors, top-down constraints are incorporated, which prevent unrealistic forecasts when the data are aggregated. Third, the hierarchical loss function allows for joint optimization of accurate forecasts at both the granular and aggregate levels. This results in more robust and calibrated forecasts across the hierarchies. Finally, the relative weighting hyperparameter provides a tuning mechanism to control the trade-off between optimizing for granular level forecasts and aggregate level forecasts. This adds flexibility in adapting hierarchical reconciliation to different use cases and requirements.

% === IV.  ========================================
% =================================================================================
\section{Experiments and results}

\subsection{Datasets and performance metrics}
\label{dataset_testing_c2}
\begin{enumerate}

 	\item \textbf{Favorita dataset:} This dataset \cite{kaggle_2017} is from a sales forecasting competition conducted in Kaggle. The dataset contains the sales history of grocery items from January 2013 to August 2017,  with information on promotions and metadata on items and stores. We generated a geographic hierarchy based on stores, cities, and states. It contains 4500 nodes with 4 levels of aggregation.  Time steps 1681 to 1687 constitute a test window of length 7. Time steps 1 to 1686 were used for the training and validation.

  	% \item \textbf{RE-Europe dataset:} This dataset \cite{jensen_desevin_greiner_pinson_2017} contained Electricity consumption records measured in kWh from 2012 to 2014 for 1494 nodes across Europe recorded hourly with 2 levels of aggregation.

     \item \textbf{The Australian Tourism dataset:} This dataset contains monthly data on domestic tourist counts in Australia. It covers 7 states and territories, each further divided into regions, sub-regions, and visit types. The dataset contained approximately 500 nodes and spanned 230 time steps. Time steps 1 to 221 were used for training and validation. The test metrics were computed in steps 222 to 228.
  
\item \textbf{M5 dataset:} This dataset \cite{kaggle_2020} originates from a sales forecasting competition conducted on Kaggle containing historic product sales data from 10 Walmart stores in three US states, along with the product hierarchy and store information. Following \citet{paria2021hierarchically}, we used only the product hierarchy. The dataset contains 3000 nodes with 4 levels of aggregation. The time steps 1907 to 1913 constitute a test window of length 7. Time steps 1 to 1906 were used for training and validation.
 
\end{enumerate}

For performance evaluation, we used absolute errors, as the evaluation was done in comparison with median predictions across methods:

\textbf{Weighted Absolute Percentage Error (WAPE)}: This metric measures the average percentage error between the true and predicted values, weighted by the magnitude of the true values. The formula for WAPE is as follows:

\begin{equation}
\label{equ:WAPE}
\mathrm{WAPE} = \frac{\sum_{(i, \tau) \in \Omega_{Test}} \left| Y_{(i, \tau)} - \hat{Y}_{(i, \tau)} \right| \cdot |Y_{(i, \tau)}|}{\sum_{(i, \tau) \in \Omega_{Test}} |Y_{(i, \tau)}|}
\end{equation}

\textbf{Mean Absolute Scaled Error (MASE)}: This metric is similar to MAE, but it is scaled by naive error. The formula for MASE is as follows:

\begin{equation}
\label{equ:MASE}
\mathrm{MASE} = \frac{ \frac{1}{\hat{N}} \sum^{\hat{N}}_{(i, \tau) \in \Omega_{Test}}  (|Y_{(i, \tau)} - \hat{Y}_{(i, \tau)}) |}{ \frac{1}{N} \sum^N_{(i, t) \in \Omega_{Train}} (|Y_{(i, t)} - \tilde{Y}_{(i, t)} |)}
\end{equation}
Here, $\tilde{Y}_{(i, t)}$ refers to the naive forecast, $Y_{(i, t)}$ represents actuals and $\hat{Y}_{(i, t)}$ represents the model forecasts.

The hyperparameters for the model were chosen following the same parameters as those published in the original literature. This ensured that the results of our study were comparable to the results of previous studies.

\subsection{Compared Methods}

As we focus in this study on point forecasts but some of the comparison methods provide probabilistic forecasts, we use their forecast of the median, that is, P50 percentile in those cases.  

\paragraph{Baseline and Current State-Of-The-Art Methods}
\begin{itemize}
    \item \textbf{Fedformer:} The FEDformer combines the advantages of transformers and seasonal-trend decomposition methods by first decomposing the time series into seasonal and trend components to extract global information. Subsequently, a frequency-enhanced transformer was used to capture a more detailed structure.
	\item \textbf{RNN-GRU:} It is a type of recurrent neural network that is used to process sequential data.
	\item \textbf{ARIMA:} ARIMA is a statistical model that uses past values of a time series to predict future values. It is made up of three components: Autoregressive (AR) terms, Moving average (MA) terms, and Integration (I).
    \item \textbf{ETS:} Exponential smoothing is a forecasting method that uses weighted averages of past observations to predict future values.
	\item \textbf{DPMN:} Please refer to Section \ref{DPMN}.
    \item \textbf{NBeats-SHARQ:} NBeats-SHARQ is a hybrid model that has double residual stacks of fully connected layers which are then combined with SHARQ.
    \item \textbf{Hier-E2E:} Please refer to Section \ref{HierE2E}.
	\item \textbf{HIRED:} Please refer to Section \ref{HIRED}.
    \item \textbf{PROFHIT:} Please refer to Section \ref{PROFHIT}. 
    \end{itemize}
    
\paragraph{Proposed Hierarchical End-to-End Graph Neural Network Methods}
 DeepHGNN refers to the stacking of the Hierarchical Aggregation Block to the base model and optimizing end-to-end with hierarchical structure as their adjacency matrix.
\begin{itemize}
	\item \textbf{DeepHGNN-DCRNN:} We used DCRNN as the base model, as outlined in Section \ref{DCRNN}.
    \item \textbf{DeepHGNN-MTGNN:} The proposed model variant using MTGNN, please refer to Section \ref{MTGNN} for details.
	\item \textbf{DeepHGNN-ADLGNN:} The proposed model variant using ADLGNN, please refer to Section \ref{ADLGNN-oursc2} for details.
    \item \textbf{DeepHGNN-MTGODE:} The proposed model variant using MTGODE, please refer to Section \ref{MTGODE} for details.
	\item \textbf{DeepHGNN-ASTGCN:} The proposed model variant using ASTGCN, please refer to Section \ref{ASTGCN} for details.
    \item \textbf{DeepHGNN-GRAMODE:}  The proposed model variant using GRAMODE, please refer to Section \ref{GRAMODE} for details.
    \item \textbf{DeepHGNN-SpecTGNN:} The proposed model variant using SpecTGNN, please refer to Section \ref{SpecTGNN} for details.
\end{itemize}

 \begin{table*}
	\caption{Comparison of state-of-the-art methods. (\textit{BU refers to bottom-up reconciliation, TD refers to top-down reconciliation, ERM refers to empirical risk minimizer reconciliation and MinT refers to Minimum trace reconciliation \cite{wickramasuriya2018optimal}.})} 
	\label{results-table_c2}
	\centering
		\begin{tabular}{lll|ll|ll}

     Model     &\multicolumn{2}{c}{M5}&  \multicolumn{2}{c}{Favorita}&  \multicolumn{2}{c}{Tourism-L}\\
 & WAPE& MASE&  WAPE&MASE& WAPE&MASE\\

\hline
 
 \multicolumn{7}{l}{\textbf{Baseline Models}}\\
 \hline
			
    %  baselines
          Fedformer (incoherent)  &0.1354& 0.7040&  0.1863&0.8569
&  0.2441    
&0.1323
\\
           Fedformer-BU       &0.1407&  0.6976
&  0.1870&0.8915
&  0.2446 
&0.1367
\\
          Fedformer-TD       &0.1361&  0.6727
&  0.2368&1.2032
&  0.2973   
&0.2184
\\
          Fedformer-ERM       &0.1353&  0.6590&  0.1786&0.7523
&  0.2421   
&0.1177
\\
          Fedformer-MinT       &0.1392&  0.6840&  0.1892&0.8921
&  0.2435   
&0.1216
\\
          RNN (incoherent)  &0.1265&  0.6504
&  0.1843&0.8370&  0.2443   
&0.1339
\\
           RNN-BU       &0.1266& 0.7021
&  0.1823&0.8341
&  0.2397
&0.1129
\\
          RNN-TD       &0.1302& 0.7025
&  0.1942&0.9695
&  0.2449  
&0.1373
\\
          RNN-ERM       &0.1513& 0.7296
&  0.2272&1.0311
&  0.2393
&0.1102
\\
          RNN-MinT       &0.1271& 0.7024
&  0.1863&0.8536
&  0.2427 
&0.1194
\\
          ARIMA-ERM  &0.1687& 0.7520&  0.1812&0.8104
&  0.2418  
&0.1186
\\
          ETS-ERM  &0.1692& 0.7426
&  0.1803&0.8386
&  0.2402
&0.1151
\\ 
\hline
 \multicolumn{7}{l}{\textbf{End-to-End SOTA Hierarchical Forecasting Methods}}\\
\hline
    % state-of-the-art
        DPMN   &0.1235&    0.6455
&  0.1796&0.7478
&  0.2370
&0.1036
\\ % dirprop
         NBeats-SHARQ   &0.1535&  0.6870&  0.1854&0.8445
&  0.2399
&0.1119
\\
          Hier-E2E    &0.1790& 0.7752
&  0.1814&0.8076
&  0.2403 
&0.1159
\\
          HIRED      &0.1337&  0.6460&  0.1861&0.8495
&  0.2412
&0.1162
\\ % 
          PROFHIT      &0.1263& 0.7042
&  0.1809&0.8191
&  0.2385
&0.1076
\\
\hline
 \multicolumn{7}{l}{\textbf{Hierarchical Graph Neural Networks (Ours)}}\\
\hline
    % Graph models
         DeepHGNN-DCRNN     &0.1347&  0.6533
&  0.1829&0.8320&  0.2398  
&0.1116
\\
         DeepHGNN-MTGNN    &0.1231&  0.6259
&  0.1790&0.7450&  0.2374 
&0.1039
\\
        DeepHGNN-ADLGNN    &0.1224& 0.7005
&  0.1732&0.7068
&  0.2377
&0.1076
\\
        DeepHGNN-MTGODE    &0.1153&  0.6089
&  0.1714&0.6846
&  0.2322 
&0.0977
\\
        DeepHGNN-ASTGCN    &0.1327&  0.6443
&  0.1795&0.7532
&  0.2368
&0.1023
\\
        DeepHGNN-GRAMODE    &0.1163&  0.6275
&  0.1708&\textbf{0.6118}&  0.2346
&0.0990\\
        DeepHGNN-SpecTGNN   &\textbf{0.1149}&  \textbf{0.6060}&  \textbf{0.1701}&0.6121&  \textbf{0.2318}&\textbf{0.0905}\\

			\hline

		\end{tabular}
\end{table*}

\begin{figure*}[htb]
	\hspace{-3cm}
 \begin{tikzpicture}[
  treatment line/.style={rounded corners=1.5pt, line cap=round, shorten >=1pt},
  treatment label/.style={font=\small},
  group line/.style={ultra thick},
]

\begin{axis}[
  clip={false},
  axis x line={center},
  axis y line={none},
  axis line style={-},
  xmin={1},
  ymax={0},
  scale only axis={false},
  width={0.9\textwidth},
  ticklabel style={anchor=south, yshift=1.3*\pgfkeysvalueof{/pgfplots/major tick length}, font=\small},
  every tick/.style={draw=black},
  major tick style={yshift=.5*\pgfkeysvalueof{/pgfplots/major tick length}},
  minor tick style={yshift=.5*\pgfkeysvalueof{/pgfplots/minor tick length}},
  title style={yshift=\baselineskip},
  xmax={24},
  ymin={-13.5},
  height={12\baselineskip},
]

\draw[treatment line] ([yshift=-2pt] axis cs:1.25, 0) |- (axis cs:-0.75, -2.0)
  node[treatment label, anchor=east] {DeepHGNN-SpecTGNN};
\draw[treatment line] ([yshift=-2pt] axis cs:2.25, 0) |- (axis cs:-0.75, -3.0)
  node[treatment label, anchor=east] {DeepHGNN-GRAMODE};
\draw[treatment line] ([yshift=-2pt] axis cs:2.5, 0) |- (axis cs:-0.75, -4.0)
  node[treatment label, anchor=east] {DeepHGNN-MTGODE};
\draw[treatment line] ([yshift=-2pt] axis cs:4.75, 0) |- (axis cs:-0.75, -5.0)
  node[treatment label, anchor=east] {DeepHGNN-ADLGNN};
\draw[treatment line] ([yshift=-2pt] axis cs:5.75, 0) |- (axis cs:-0.75, -6.0)
  node[treatment label, anchor=east] {DeepHGNN-MTGNN};
\draw[treatment line] ([yshift=-2pt] axis cs:6.0, 0) |- (axis cs:-0.75, -7.0)
  node[treatment label, anchor=east] {DPMN};
\draw[treatment line] ([yshift=-2pt] axis cs:8.0, 0) |- (axis cs:-0.75, -8.0)
  node[treatment label, anchor=east] {HGNN-ASTGCN};
\draw[treatment line] ([yshift=-2pt] axis cs:8.0, 0) |- (axis cs:-0.75, -9.0)
  node[treatment label, anchor=east] {PROFHIT};
\draw[treatment line] ([yshift=-2pt] axis cs:11.5, 0) |- (axis cs:-0.75, -10.0)
  node[treatment label, anchor=east] {RNN-BU};
\draw[treatment line] ([yshift=-2pt] axis cs:12.5, 0) |- (axis cs:-0.75, -11.0)
  node[treatment label, anchor=east] {Fedformer-ERM};
\draw[treatment line] ([yshift=-2pt] axis cs:14.0, 0) |- (axis cs:-0.75, -12.0)
  node[treatment label, anchor=east] {RNN(incoherent)};
\draw[treatment line] ([yshift=-2pt] axis cs:14.0, 0) |- (axis cs:-0.75, -13.0)
  node[treatment label, anchor=east] {HIRED};
\draw[treatment line] ([yshift=-2pt] axis cs:14.125, 0) |- (axis cs:23.5, -13.0)
  node[treatment label, anchor=west] {DeepHGNN-DCRNN};
\draw[treatment line] ([yshift=-2pt] axis cs:14.5, 0) |- (axis cs:23.5, -12.0)
  node[treatment label, anchor=west] {Hier-E2E};
\draw[treatment line] ([yshift=-2pt] axis cs:15.5, 0) |- (axis cs:23.5, -11.0)
  node[treatment label, anchor=west] {RNN-ERM};
\draw[treatment line] ([yshift=-2pt] axis cs:15.625, 0) |- (axis cs:23.5, -10.0)
  node[treatment label, anchor=west] {RNN-MinT};
\draw[treatment line] ([yshift=-2pt] axis cs:17.25, 0) |- (axis cs:23.5, -9.0)
  node[treatment label, anchor=west] {ETS-ERM};
\draw[treatment line] ([yshift=-2pt] axis cs:17.375, 0) |- (axis cs:23.5, -8.0)
  node[treatment label, anchor=west] {Fedformer(incoherent)};
\draw[treatment line] ([yshift=-2pt] axis cs:17.75, 0) |- (axis cs:23.5, -7.0)
  node[treatment label, anchor=west] {NBeats-SHARQ};
\draw[treatment line] ([yshift=-2pt] axis cs:18.0, 0) |- (axis cs:23.5, -6.0)
  node[treatment label, anchor=west] {ARIMA-ERM};
\draw[treatment line] ([yshift=-2pt] axis cs:18.75, 0) |- (axis cs:23.5, -5.0)
  node[treatment label, anchor=west] {RNN-TD};
\draw[treatment line] ([yshift=-2pt] axis cs:18.875, 0) |- (axis cs:23.5, -4.0)
  node[treatment label, anchor=west] {Fedformer-MinT};
\draw[treatment line] ([yshift=-2pt] axis cs:20.25, 0) |- (axis cs:23.5, -3.0)
  node[treatment label, anchor=west] {Fedformer-BU};
\draw[treatment line] ([yshift=-2pt] axis cs:21.5, 0) |- (axis cs:23.5, -2.0)
  node[treatment label, anchor=west] {Fedformer-TD};
\draw[group line] (axis cs:6.0, -4.666666666666667) -- (axis cs:8.0, -4.666666666666667);
\draw[group line] (axis cs:5.75, -4.0) -- (axis cs:6.0, -4.0);
\draw[group line] (axis cs:8.0, -5.333333333333333) -- (axis cs:14.0, -5.333333333333333);
\draw[group line] (axis cs:2.25, -2.0) -- (axis cs:2.5, -2.0);
\draw[group line] (axis cs:1.25, -1.3333333333333333) -- (axis cs:2.25, -1.3333333333333333);
\draw[group line] (axis cs:14.5, -2.0) -- (axis cs:20.25, -2.0);
\draw[group line] (axis cs:11.5, -6.666666666666667) -- (axis cs:15.5, -6.666666666666667);
\draw[group line] (axis cs:12.5, -3.3333333333333335) -- (axis cs:18.75, -3.3333333333333335);
\draw[group line] (axis cs:14.5, -2.2) -- (axis cs:20.25, -2.2);
\draw[group line] (axis cs:15.5, -1.3333333333333333) -- (axis cs:21.5, -1.3333333333333333);
\draw[group line] (axis cs:15.5, -1.5333333333333332) -- (axis cs:21.5, -1.5333333333333332);

\end{axis}
\end{tikzpicture}
	\caption{The critical difference diagram shows the mean ranks of each model evaluated.}
	\label{fig:cdDiach2}
\end{figure*}
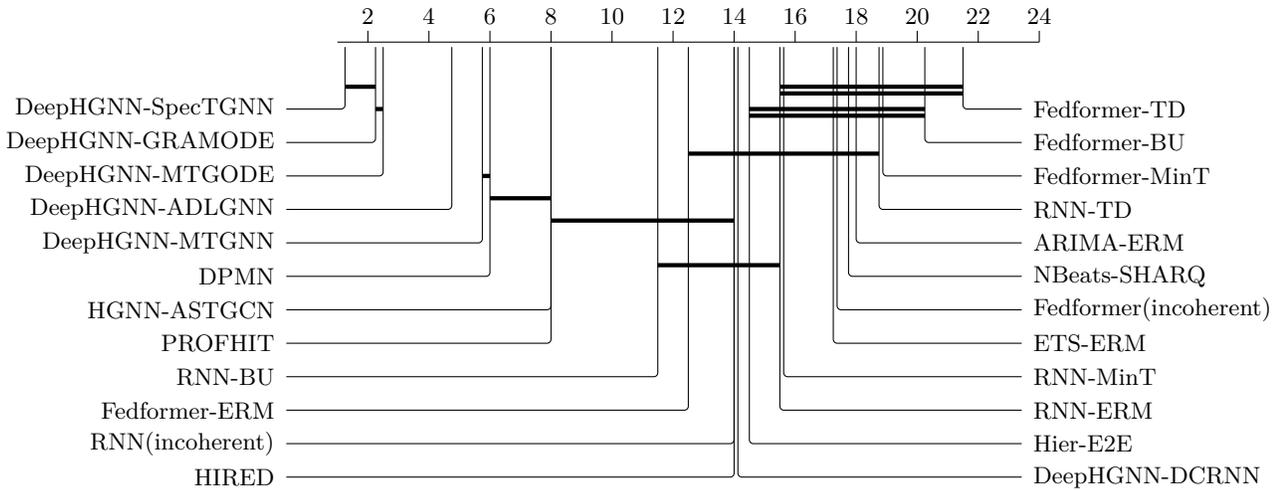

\subsection{Experimental Setup}
\label{hyperparameters-setup_c2}
 The training datasets were divided into a training set (80\%) and a validation set (20\%) and the last window was used for testing. The models were trained using the hyperparameters proposed in the original implementation. An early stopping mechanism was implemented to stop training when no improvement was observed in the validation metrics for over four epochs.

\section{Result \& Discussion}

Table \ref{results-table_c2} shows the experimental results of the proposed methods along with other existing methods. From the results, we can observe that most of our proposed DeepHGNN model variants consistently outperformed the existing models on both MASE and WAPE. The DeepHGNN-specTGNN is the best performing model overall. 

The baseline models, including various configurations of the Fedformer and RNN, along with traditional methods like ARIMA and ETS (both with ERM), exhibited varying degrees of performance across datasets. RNN models generally outperformed Fedformer but were still surpassed by other advanced methods. Notably, traditional models such as ARIMA and ETS with ERM reconciliation, revealed higher error rates, highlighting the limitations of conventional approaches in this context. Among the end-to-end SOTA hierarchical forecasting methods, DPMN has emerged as a strong contender. Other models such as NBeats-SHARQ, Hier-E2E, HIRED, and PROFHIT, while innovative in their approach, demonstrated moderate to high error rates. The significant relative gains shown in Table \ref{tab:relative gain_c2}  not only underline the effectiveness of DeepHGNNs but also highlight the potential of graph neural network techniques in hierarchical forecasting scenarios.

The critical difference diagram based on the WAPE in Figure \ref{fig:cdDiach2} provides a statistical comparison of the forecasting accuracy for the different models evaluated in our experiments. The models are ranked from best to worst based on their average rank of forecast errors, with lower ranks indicating better accuracy. Models connected by horizontal lines are considered not significantly different, whereas models not connected by a line have significantly different performances.

The DeepHGNN models (DeepHGNN-SpecTGNN, DeepHGNN-GRAMODE, DeepHGNN-MTGODE, DeepHGNN-ADLGNN, DeepHGNN-MTGNN, and DeepHGNN-ASTGCN) were grouped on the left side of the diagram, indicating their superior performance. Notably, the DeepHGNN-SpecTGNN is the leftmost model, reflecting its status as the top-performing model.  Models such as DPMN, PROFHIT, and others like Fedformer variations and RNN variations are grouped towards the middle and right of the diagram. This positioning indicates a lower average rank compared to the DeepHGNN models. Traditional baseline models, including those based on ARIMA and ETS, are ranked on towards the right, showing relatively poor performance. The relative distance between the DeepHGNN models and the rest of the models imply that the improvements are not just incremental; they are substantial and statistically significant, demonstrating the generalizability of the approach. The absence of connecting lines between the top performing DeepHGNN models (DeepHGNN-SpecTGNN, DeepHGNN-GRAMODE, DeepHGNN-MTGODE and DeepHGNN-ADLGNN) and the rest indicates a robust statistically significant difference in performance from both the baselines and SOTA models.

\begin{table}
    \centering
    \begin{tabular}{ccccc}
    \hline
          Model &   M5 &   Favorita &    Tourism-L &     Overall \\
          \hline
  DeepHGNN-MTGODE & 6.640\% & 4.566\% &  2.025\% & 3.925\% \\ 
 DeepHGNN-GRAMODE & 5.830\% & 4.900\% &  1.012\% & 3.406\% \\
DeepHGNN-SpecTGNN & 6.964\% & 5.290\% &  2.194\% & 4.314\% \\
    \end{tabular}
    \caption{Comparative Analysis: Demonstrating the Percentage Reduction in WAPE Error by the Top 3 DeepHGNN Models Against Existing State-of-the-Art Methods}
    \label{tab:relative gain_c2}
\end{table}

\section{Conclusions}

In this study, we proposed a hierarchical graph neural network (DeepHGNN) approach to model hierarchical dependencies in time series data. Our experiments demonstrate that incorporating the hierarchical structure and reconciliations through graph neural networks significantly improves forecasting accuracy compared to existing methods. The critical difference analysis verified that DeepHGNNs achieved statistically lower errors than popular benchmarks including RNNs, LSTMs, and temporal convolution networks across various datasets. By jointly modeling the interactions within and across levels, DeepHGNNs provide more accurate and calibrated forecasts. DeepHGNNs enable feature sharing and reconciliation regularization, which are lacking in traditional forecasting models.

Overall, our results highlight the benefits of explicitly encoding domain-specific hierarchical constraints using graph neural networks. This is evidence that hierarchical graph neural networks are promising modeling technique for forecasting problems involving complex hierarchical time series data. 

A flexible DeepHGNN framework can encode diverse reconciliation constraints and dependencies.
Although DeepHGNNs demonstrate significant improvements in forecasting accuracy over traditional methods, there remain some limitations and open challenges that provide avenues for further research. A key limitation is that computational and memory demands scale rapidly for modeling very large hierarchies with thousands of time series, limiting the applicability for big complex datasets that present opportunities for advancement through future research.

\section*{Acknowledgment}

This research was supported by the Facebook Statistics for Improving Insights and Decisions research award, Monash University Graduate Research funding, and the MASSIVE - High performance computing facility, Australia. Christoph Bergmeir is currently supported by a María Zambrano (Senior) Fellowship that is funded by the Spanish Ministry of Universities and Next Generation funds from the European Union. 

\bibliography{Bibliography}

\end{document}